\documentclass{article} %
\usepackage{iclr2027_conference,times}

\usepackage{amsmath,amsfonts,bm}

\def\eqref#1{equation~\ref{#1}}

\def\1{\bm{1}}

\DeclareMathAlphabet{\mathsfit}{\encodingdefault}{\sfdefault}{m}{sl}
\SetMathAlphabet{\mathsfit}{bold}{\encodingdefault}{\sfdefault}{bx}{n}

\usepackage{hyperref}

\hypersetup{
  colorlinks=true,
  linkcolor=red,     
  urlcolor=magenta,
  citecolor=teal!80!black
}

\usepackage{url}
\usepackage{graphicx}
\usepackage{booktabs}
\usepackage{adjustbox}
\usepackage{amssymb}
\usepackage{multirow}
\usepackage{subcaption}
\usepackage{enumitem}
\usepackage{capt-of}
\usepackage[table]{xcolor}
\usepackage{microtype}
\def\modelname{BUMP}

\def\Title{Body-driven Unified Multi-object Physics for HOI synthesis}
\def\paperTitle{\textit{\modelname}: \Title}

\def\authorBlock{%
    Wonjung Woo, Hyeonwoo Kim, Hanbyul Joo \\
    Seoul National University\\
    \texttt{\{agnes2327, hwkim408, hbjoo\}@snu.ac.kr} \\
    \url{https://wonjungwoo.github.io/BUMP/}
}

\title{\paperTitle}

\author{\authorBlock}

\iclrfinalcopy %
\begin{document}

\maketitle
\fancyhead{}

\begin{abstract}
Generating human-object interactions (HOIs) with multiple objects is challenging because the resulting motions arise from different underlying causes.
Human motion reflects semantic intent, intentionally manipulated objects must remain coordinated with the human, while surrounding objects move primarily as physical consequences of contact and collision.
We present \modelname, a text-driven framework for multi-object HOI generation based on a simple principle: learn intentional motion and simulate physical consequences.
Our method first generates full-body human motion from text and scene context, then predicts trajectories for a small number of intentionally manipulated \emph{driver} objects conditioned on the human motion.
We then obtain \emph{responder} object motions through physics simulation, allowing the framework to handle varying numbers and configurations of surrounding objects without expanding the learned trajectory output space.
To address the scarcity of paired full-body multi-object HOI data, we combine heterogeneous supervision from human-only motion, full-body HOI, and hand-centric manipulation data augmented with plausible body motion.
We further introduce outcome-conditioned interaction editing, which infers a local modification of the causal driver object trajectory from a desired responder object behavior without requiring an explicit target driver object trajectory.
Experiments demonstrate strong performance on both single-object and multi-object HOI benchmarks.
Qualitative results showcase diverse scenarios involving inter-object relationships and edits, highlighting the advantage of our unified model formulation.
\end{abstract}

\section{Introduction}
\label{sec:intro}

Human activities in the real world rarely involve a single object in isolation.
Everyday behaviors such as preparing food or playing sports often involve intentionally manipulating one or two objects while interacting with several surrounding objects.
While most HOI generation methods focus on a single manipulated object
~\citep{OMOMO,CHOIS,InterDiff,HOIDiff,InterDreamer,HOIfHLI},
recent work has begun to address multi-object interactions
~\citep{HIMO,lu2025humoto,PAMotion,Dex2HOI}.
However, explicitly generating every interacting object couples human motion,
multiple object trajectories, contacts, and inter-object relationships within
the learned generation problem.

We argue that these sources of complexity need not be modeled symmetrically.
The semantic structure of an interaction is often expressed primarily by
human motion and intentionally manipulated objects.
For example, swinging a baseball bat is defined by the human and bat motions, while the subsequent motion of a struck ball, or other objects displaced by that ball, is largely determined by contact dynamics.
This motivates a simple principle: \emph{learn intentional motion and simulate physical consequences}.

Based on this principle, we propose \modelname{}, a text-conditioned framework that decomposes an interaction into the human, intentionally manipulated \emph{driver objects}, and physically induced \emph{responder objects}.
We first generate full-body human motion from text and scene context, then generate a small number of driver object trajectories conditioned on the human.
Responder object trajectories are obtained through physics simulation from contacts with the human, driver objects, or other responder objects.
Surrounding objects can therefore condition the learned motion without becoming trajectory outputs, decoupling the learned trajectory space from the number of responder objects.

To address limited full-body multi-object HOI supervision, we combine human-only motion, captured full-body HOI, and fine-grained manipulation data.
Hand-centric and partial-body sequences are lifted to a common full-body representation through offline fitting and retargeting, with reconstructed body regions treated as weaker supervision.

This decomposition also enables \emph{outcome-conditioned interaction editing}: given a desired responder object behavior, we locally modify the causal driver object motion and physically replay the interaction to realize the requested outcome.

\setlength{\dblfloatsep}{2pt}
\begin{figure*}[t]
    \centering
    \includegraphics[width=\textwidth]{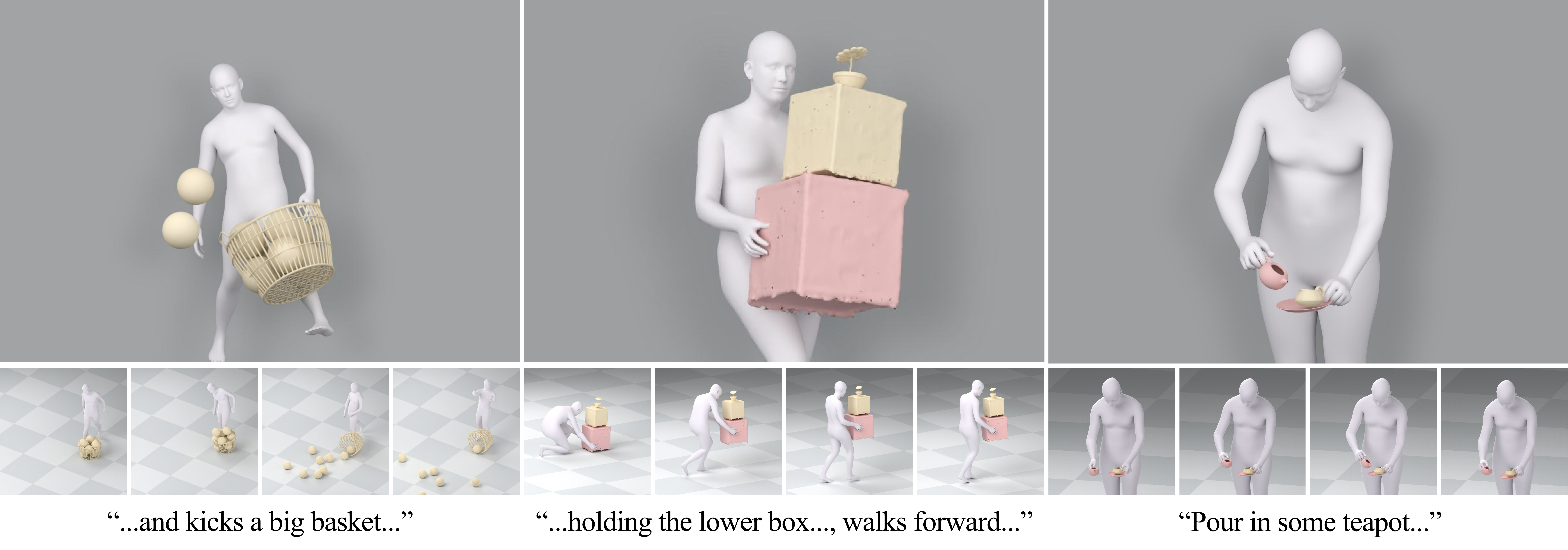}
    \caption{
    \textbf{Overview of \modelname{}.}
    Given a text prompt and an initial scene, \modelname{} generates intentional
    human and driver object motion while obtaining physically induced responder object
    motion through simulation.
    }
    \label{fig:teaser}
\end{figure*}

In summary, our main contributions are:
\begin{itemize}[leftmargin=1.2em,topsep=2pt,itemsep=1pt,parsep=0pt]
\item We propose \modelname{}, a \textbf{factorized multi-object HOI framework} that learns human and driver object motion and simulates responder object motion, decoupling the learned trajectory space from the number of responder objects.
\item We introduce a \textbf{heterogeneous full-body interaction data construction strategy} that lifts fine-grained hand-centric and partial-body manipulation data to a common full-body representation, providing additional dexterous and multi-object supervision.
\item We introduce \textbf{outcome-conditioned interaction editing} that locally modifies driver object motion to achieve a desired responder object behavior without an explicit target driver object trajectory.
\end{itemize}

Code and models will be released.

\section{Related Work}
\label{sec:related}

\noindent\textbf{Human-Object Interaction Modeling.}
Prior work has modeled human-object affordances, contacts, and geometric
relationships from 3D observations~\citep{3DAffordanceNet,IAGNet,LEMON,DECO,ComA},
providing useful priors for interaction locations.
Full-body HOI generation methods such as OMOMO~\citep{OMOMO},
CHOIS~\citep{CHOIS}, InterDiff~\citep{InterDiff}, HOIDiff~\citep{HOIDiff},
AffordMotion~\citep{AffordMotion}, TesMo~\citep{TesMo}, and
InterDreamer~\citep{InterDreamer} further synthesize full-body human motion for object and scene interactions.
Recent approaches improve generalization or physical plausibility, e.g.,
DAViD~\citep{DAViD} adapts a pretrained human-motion prior to HOI, while
HOIfHLI~\citep{HOIfHLI} combines kinematic generation with physics-based
tracking.
Methods that jointly generate human and object motion primarily focus on a single manipulated object.

\noindent\textbf{Multi-Object HOI Generation.}
Recent work extends HOI synthesis to multiple objects.
HIMO~\citep{HIMO} generates full-body two- and three-object interactions with
a dual-branch diffusion model, while Dex2HOI~\citep{Dex2HOI} coordinates
multiple object-motion streams for dexterous one- and two-object manipulation.
PAMotion~\citep{PAMotion} further introduces physics-aware objectives for
joint generation of human and multi-object trajectories.
In these approaches, interacting object motions remain explicit outputs of
learned generative models.
Our work instead separates intentionally manipulated driver objects from
physically induced responder objects, learning only human and driver object
motion while obtaining responder object trajectories through simulation.
This allows additional objects without expanding the learned trajectory output space.

\noindent\textbf{HOI Data and Heterogeneous Supervision.}
Full-body multi-object HOI data remains limited.
HIMO~\citep{HIMO}, HUMOTO~\citep{lu2025humoto}, and ParaHome~\citep{ParaHome}
capture multi-object interactions, while GRAB~\citep{GRAB},
InterCap~\citep{InterCap}, BEHAVE~\citep{BEHAVE}, ARCTIC~\citep{ARCTIC},
and InterAct~\citep{InterAct} provide complementary HOI supervision.
We additionally exploit broader human-motion data and fine-grained
manipulation datasets such as TACO~\citep{TACO} and OakInk2~\citep{OakInk2},
lifting heterogeneous sources into a common full-body representation.

\noindent\textbf{Physics-Based Interaction and Editing.}
Existing physics-aware HOI methods use physics to regularize, refine, or track
motions produced by learned models~\citep{PAMotion,HOIfHLI}.
In contrast, our simulator directly produces responder object trajectories,
making physics part of the generative process itself.
We further exploit this causal structure for outcome-conditioned editing by
modifying the causal driver object motion from a desired responder object outcome rather
than requiring a prescribed object trajectory.

\section{Method}

\begin{figure}[t]
    \centering
    \includegraphics[width=\linewidth]{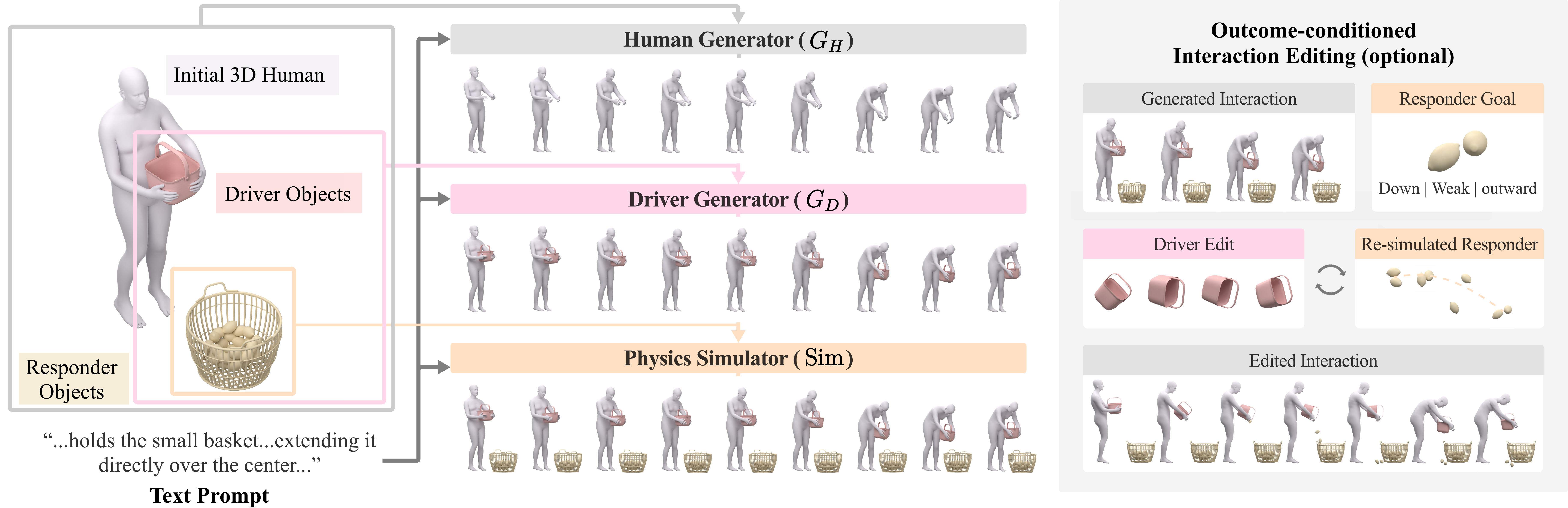}
    \caption{
    \textbf{Overview of the generation and editing pipeline.}
    We generate human motion, predict hand-driver object contacts, and generate driver object trajectories conditioned on both. Physics simulation produces responder object motions.
    For outcome-conditioned editing, we locally modify the driver object trajectory, validate the resulting responder object motion through simulation, and align the human motion with the edited driver object while preserving the original motion.
    }
    \label{fig:method_overview}
\end{figure}

Given a text prompt and an initial scene, \modelname{} generates multi-object HOIs by learning human and driver object motion and simulating responder object motion. \emph{Driver objects} are those intentionally manipulated by the human, whereas \emph{responder objects} move as physical consequences of contact and collision involving the human, driver objects, or other responder objects.
Formally, given a text prompt $\ell$, an initial human pose $\mathbf h_1$, an initial scene $\mathcal O^0$ containing object geometries and poses, and responder object initialization $\mathcal S_R^0$, \modelname{} produces the human trajectory $\mathcal H$, driver object trajectories $\mathcal D$, and responder object trajectories $\mathcal R$:
\begin{equation}
(\mathcal H,\mathcal D,\mathcal R)
=
\mathrm{BUMP}(\ell,\mathbf h_1,\mathcal O^0, \mathcal S_R^0).
\end{equation}
The pipeline consists of three stages that follow the causal order:
\begin{align}
\mathcal H &= G_H(\ell,\mathbf h_1,\mathcal O^0), \label{eq:stage1}\\
\mathcal D &= G_D(\mathcal H,\mathcal C,\ell,\mathcal O^0), \label{eq:stage2}\\
\mathcal R &= \mathrm{Sim}(\mathcal H,\mathcal D,\mathcal S_R^0;\psi). \label{eq:stage3}\\
\mathcal C &= P_C(\mathbf Z_H,\mathcal Q_D), \label{eq:contact}
\end{align}
$G_H$ and $G_D$ are diffusion transformers that generate human and driver object motion, respectively. Their architectures follow prior work~\citep{rempe2026kimodo, CHOIS}, with input and output layers adapted to our motion representation. 
$\mathrm{Sim}$ is a rigid-body physics simulator with physical parameters $\psi$.
The responder object initialization $\mathcal S_R^0$ specifies object geometries, initial poses, and linear and angular velocities.
The auxiliary contact head $P_C$ predicts contacts between hands and driver objects $\mathcal C$ from the frame-wise hidden features $\mathbf Z_H$ of $G_H$ and the selected driver object queries $\mathcal Q_D$.
See Figure~\ref{fig:method_overview} for an overview.

\subsection{Problem Setup and Notation}

\noindent\textbf{Human.}
The human trajectory is a sequence of $T$ frames, $\mathcal H=\{\mathbf h_t\}_{t=1}^{T}$, where each frame is
\begin{equation}
\mathbf h_t=
\left[
\mathbf p^{\mathrm{root}}_t,\;
\boldsymbol\theta^{\mathrm{body}}_t,\;
\boldsymbol\theta^{\mathrm{hand}}_t
\right]
\in\mathbb R^{3 + 22 \times 6 + 30 \times 6}.
\end{equation}
Here $\mathbf p^{\mathrm{root}}_t$ is the root translation, $\boldsymbol\theta^{\mathrm{body}}_t$ stacks the global rotations of the 22 body joints, and $\boldsymbol\theta^{\mathrm{hand}}_t$ stacks the local rotations of the 30 finger joints (15 per hand). All rotations use the 6D representation~\citep{zhou2019continuity}. The world coordinate frame is defined by the root joint frame of the initial pose $\mathbf h_1$.
We denote the final frame-wise hidden features of the human generator by
$\mathbf Z_H=\{\mathbf z_{H,t}\}_{t=1}^{T}$, where
$\mathbf z_{H,t}\in\mathbb R^{1024}$.

\noindent\textbf{Input scene and object roles.}
The scene contains $N$ objects, $\mathcal O^0=\{\mathbf o_i\}_{i=1}^{N}$, each represented as 
\begin{equation}
\mathbf o_i=
\left[
\mathbf g_i,\;
\mathbf p^{0}_i,\;
\mathbf R^{0}_i
\right]
\in\mathbb R^{256 + 3 + 9},
\end{equation}
where $\mathbf g_i \in\mathbb R^{256}$ is an encoded feature of a
1,024-point Basis Point Set (BPS) representation~\citep{prokudin2019efficient, CHOIS}, $\mathbf p^{0}_i$ is the initial translation, and $\mathbf R^{0}_i$ is the flattened initial rotation matrix.
Objects are partitioned into driver objects $\mathcal O_D$ and responder objects $\mathcal O_R$, with at most two driver objects per scene.
Training roles are assigned automatically from hand proximity and object motion (Sec.~\ref{sec:data}). Inference roles are given with the scene.
The model uses four object slots, retaining all driver objects and filling the remaining slots with objects closest to the human. Additional objects are simulated as responder objects. Unused slots are masked in all stages.

For contact prediction, $\mathcal Q_D$ contains one 3088-D query per selected driver object, combining its raw $1024\times3$ BPS descriptor, 12-D initial pose,
and 4-D physical-slot one-hot encoding.

\noindent\textbf{Output trajectories and contacts.}
The pose of driver object $i$ at frame $t$ is
$\mathbf d^{\,i}_t=(\mathbf p^{i}_t,\mathbf R^{i}_t)\in SE(3)$, and
$\mathcal D=\{\mathbf d^{\,i}_t\}$ denotes the driver object trajectories.
Similarly, $\mathbf r^{\,j}_t\in SE(3)$ denotes the pose of responder object
$j$, and $\mathcal R=\{\mathbf r^{\,j}_t\}$ denotes the responder object trajectories.
The contact prediction $\mathcal C$ contains, for each hand and frame, a
categorical distribution over no contact and the two driver objects.

\subsection{Stage 1: Human Motion Generation}
\label{sec:human}
We first generate the human trajectory $\mathcal H$, which serves as a fixed condition for the subsequent stages (Eq.~\ref{eq:stage1}).
$G_H$ is trained to recover the clean sequence $\mathcal H$ from its noised version, conditioned on the text $\ell$, the initial pose $\mathbf h_1$, and the scene $\mathcal O^0$. 
Each input object $\mathbf o_i$ is embedded as a separate token, giving $N \le 4$ scene tokens.

\noindent\textbf{Optional human guidance.}
For qualitative visualization, we optionally guide reverse diffusion toward
contact and force-closure grasp targets while penalizing penetration and
temporal artifacts; this guidance is not used for quantitative evaluation.

\subsection{Stage 2: Contact Prediction and Driver Object Generation}
\label{sec:driver}
Driver object motion is represented relative to the manipulating hand.
We first predict contacts between hands and driver objects $\mathcal C$
(Eq.~\ref{eq:contact}) whose hard categorical assignments determine
the hand-relative coordinate frame for driver object decoding.

\noindent\textbf{Hand-driver object contact prediction.}
The lightweight contact head $P_C$ takes as input the frame-wise final hidden features $\mathbf Z_H$ of $G_H$ and the selected driver object queries $\mathcal Q_D$ (Eq.~\ref{eq:contact}).
For each hand and frame, it predicts a categorical distribution over no contact
and the two driver objects.
With $G_H$ frozen, we train $P_C$ using 
$\mathcal L_C=\mathcal L_{\mathrm{CE}}+0.01\mathcal L_{\mathrm{temp}}$,
where $\mathcal L_{\mathrm{temp}}$ encourages temporal consistency when the
ground-truth assignment remains unchanged.

\noindent\textbf{Driver object motion generator.}
$G_D$ is conditioned on human joint positions and wrist orientations recovered from $\mathcal H$, contacts $\mathcal C$, text $\ell$, and scene $\mathcal O^0$ (Eq.~\ref{eq:stage2}).
Internally, it converts the contact probabilities into hard one-hot assignments indicating which driver object, if any, each hand contacts, and uses them for conditioning and hand-relative decoding.
For each driver object $i$ and frame $t$, it predicts
\begin{equation}
\bar{\mathbf d}^{\,i}_t=
\left[
\Delta\mathbf p^{i}_{\mathrm L,t},\;
\Delta\mathbf p^{i}_{\mathrm R,t},\;
\bar{\mathbf R}^{i}_t
\right]
\in\mathbb R^{3 + 3 + 9},
\end{equation}
consisting of translation offsets from the left and right hand anchors and an unconstrained rotation, giving a 30-dimensional output per frame for the two driver object slots.
The hard contact assignments select the hand-relative candidate used to reconstruct each driver object's position.
When both hands contact the same driver object, the two candidates are averaged. 
When neither hand contacts it, a fixed fallback hand is used.
We project $\bar{\mathbf R}^{i}_t$ onto $SO(3)$ and combine it with the decoded position to obtain the final driver object pose $\mathbf d^{i}_t$.

We train $G_D$ with the standard diffusion objective on ground-truth human motion and contact labels, then fine-tune it with an additional cubic B-spline residual loss.
At inference, generated $\mathcal H$ and the predicted $\mathcal C$ replace the ground-truth conditions, with $\mathcal H$ fixed throughout driver object sampling.

\noindent\textbf{Optional object guidance.}
For qualitative visualization, we optionally apply differentiable guidance
during driver object reverse diffusion.
At each guided step, we update the predicted clean driver object trajectory using
contact- and grasp-consistency objectives together with temporal smoothness
and penetration penalties, before computing the posterior transition.
Hard categorical contacts determine the hand-relative COM interpretation,
while soft contact probabilities modulate the strength of contact guidance.
This guidance is applied only during sampling and is not used for quantitative
benchmark results.

\subsection{Stage 3: Responder Objects by Physics Simulation}
\label{sec:responders}

\begin{figure*}[t]
    \centering
    \includegraphics[width=\textwidth]{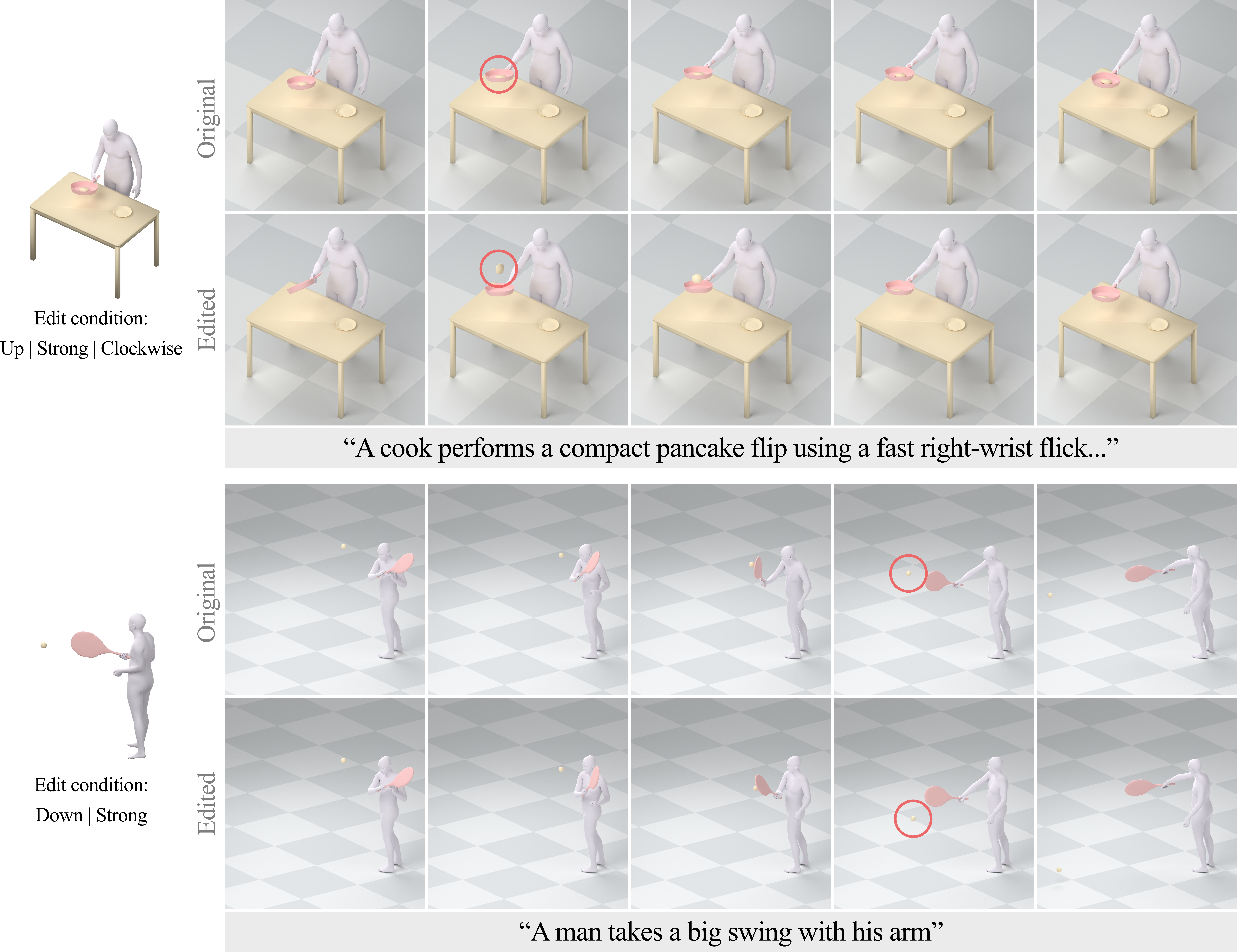}
    \caption{
    \textbf{Outcome-conditioned interaction editing.}
        We compare the original and edited interactions under different responder object-level goals.
        The edits produce the requested changes in the physical outcomes while preserving the overall interaction.
    }
    \label{fig:driver_editing}
\end{figure*}

We obtain responder object trajectories $\mathcal R$ by kinematically replaying human and driver object motion in a rigid-body physics simulator and letting responder objects react to contacts (Eq.~\ref{eq:stage3}).
We build a MuJoCo~\citep{todorov2012mujoco} scene from the object meshes and their initial poses.
Each mesh is decomposed into multiple convex collision parts with CoACD~\citep{wei2022coacd}.
Responder objects are dynamic rigid bodies with a shared uniform density and the same MJCF contact parameters, so no per-object mass or inertia measurements are required.
The generated human motion drives capsule and sphere collision proxies attached to the body, with human-driver object collisions disabled.
The simulator never modifies the prescribed human and driver object motion.

The human and driver object trajectories are defined at 30\,fps and replayed at 60\,fps using spline interpolation, with 36 simulation substeps per playback frame.

\subsection{Outcome-Conditioned Interaction Editing}
\label{sec:editing}

Given a desired responder object outcome $g_R$, we locally edit the causal driver object trajectory, evaluate the response through simulation, and retarget the human motion to the edited driver object. 
Goals specify either a motion direction and response strength, optionally with a rotation direction, or a spatial relation to preserve.

For motion goals, we map the direction category to a unit vector $\mathbf u\in\mathbb R^3$ at the interaction event.
Horizontal directions are defined relative to the human yaw, while vertical directions follow the world vertical axis.
Response strength specifies a directional speed interval $[s_{\min},s_{\max}]$, and $\alpha$ is the allowed angular deviation from $\mathbf u$.

For the responder object's post-contact velocity
$\mathbf v_r \in\mathbb R^3$, the feasible response set is
\begin{equation}
\mathcal V(g_R)
=
\left\{
\mathbf v_r \in\mathbb R^3
\;\middle|\;
s_{\min}
\le
\mathbf u^\top\mathbf v_r
\le
s_{\max},
\;
\left\|
\mathbf v_r-
(\mathbf u^\top\mathbf v_r)\mathbf u
\right\|_2
\le
\tan(\alpha)\,
\mathbf u^\top\mathbf v_r
\right\}.
\end{equation}
The constraints bound the velocity component along $\mathbf u$ and restrict its direction to a cone of
half-angle $\alpha$ around that direction.
We measure motion-goal error as the Euclidean distance from the simulated
velocity to $\mathcal V(g_R)$.
An optional rotation goal specifies an axis in the human frame at the interaction event that the responder object's post-contact angular velocity should follow. 

For relation goals, we instead measure violations of the requested relation over the simulated trajectory.
For example, support preservation constrains the responder object to remain near its initial driver object-relative pose, optionally with persistent contact.

\noindent\textbf{Local driver object correction.}
We replay the original interaction and select an event relevant to the goal.
We compute a requested responder object velocity change $\Delta\mathbf v_r$ toward $\mathcal V(g_R)$ for motion goals or from relation residuals for relation goals.
Rotation goals additionally induce rotational corrections in the same SE(3) controls.

Within a short window around the selected event, we parameterize the driver object edit by $\delta\in\mathbb R^{5\times (3+3)}$ containing five temporal controls, each with translation and rotation corrections.
We interpolate these controls with compactly supported basis functions whose value and first two derivatives
vanish at the window boundaries, preserving the trajectory outside the window.

At the selected contact, we convert $\Delta\mathbf v_r$ to an impulse, project it onto the local Coulomb friction cone, and derive a feasible driver object contact-point velocity change $\Delta\mathbf v_c$.
For small local edits, we linearize the driver object contact-point velocity with
respect to the control vector $\delta$ and solve
\begin{equation}
\delta^{*}
=
\arg\min_{\delta}
\left\|
\frac{\partial\mathbf v_c}{\partial\delta}\,\delta
-
\Delta\mathbf v_c
\right\|_2^2
+
\lambda_p\|\delta_p\|_2^2
+
\lambda_R\|\delta_R\|_2^2,
\end{equation}
where $\delta_p$ and $\delta_R$ are the translation and rotation controls.
The objective matches the requested contact-point velocity change while favoring small translation and rotation corrections.

\noindent\textbf{Simulation-guided selection and human retargeting.}
We test scaled candidate corrections through full MuJoCo replay and accumulate only those that improve the responder object objective.
The procedure is repeated until the goal is met, no improvement is found, or the iteration
budget is exhausted.

To retarget contacting limbs, we transfer the driver object's SE(3) modification to wrist targets and optimize cubic B-spline residual arm rotations.
When needed, we adjust the torso with coarser, more strongly regularized controls.
During retargeting, the edited driver object trajectory, root translation, finger articulation, and non-edited joints remain fixed.

\subsection{Heterogeneous Interaction Data Construction}
\label{sec:data}

\noindent\textbf{Data collection and unification.}
We combine human-only motion, captured full-body HOI, and fine-grained manipulation data into a common full-body representation.
Human-only motion comes from HumanML3D, Motion-X++, AIST++, BEAT2, OpenBiomechanicsProject, and LAFAN1~\citep{Guo2022HumanML3D,zhang2025motion,AISTPP,BEAT2,OBP,LAFAN1}.
Full-body interactions are obtained from OMOMO, HuMoTo, GRAB, and InterAct~\citep{OMOMO,lu2025humoto,GRAB,InterAct}, while TACO and OakInk2 provide detailed hand-object manipulation~\citep{TACO,OakInk2}.
We map all human motions to a common neutral-body model, resample all sequences to 30\,fps, and segment them into 120-frame windows.

\noindent\textbf{Object role assignment.}
We assign object roles automatically per training window using object motion and hand proximity.
Moving objects whose sampled surfaces come within 2\,cm of a hand joint in any frame are driver object candidates; others are \emph{responder objects}.
We retain at most two driver objects, selecting the highest-scoring candidates when necessary.

\noindent\textbf{Full-body fitting from incomplete observations.}
For hand-only TACO sequences, we preserve observed finger articulation and fit a neutral SMPL-X body to wrist trajectories, regularizing unobserved regions toward a neutral standing posture and encouraging temporal smoothness.
For OakInk2, we retarget the observed upper body and hands, then complete the lower body with grounded, stationary feet while limiting changes to the observed upper body.

\noindent\textbf{Canonicalization and supervision.}
For each window, human and object trajectories share a canonical coordinate frame defined by the initial human heading and horizontal root position.
Hand-object contact labels use the same 2\,cm surface-proximity criterion as driver object identification.
Dataset-specific supervision masks account for missing or reconstructed body channels; observed manipulation signals are treated as more reliable than synthetically completed body motion.

\section{Experiments}
\begin{table*}[t]
\centering
\caption{
\textbf{Quantitative comparison across single- and multi-object HOI settings.}
We report generation quality and interaction quality.
Contact F1 is reported in percentage.
Metrics are comparable only within each interaction setting.
}
\label{tab:main_baselines}

\renewcommand{\arraystretch}{1.0}
\setlength{\tabcolsep}{5.0pt}
\footnotesize
\setlength{\aboverulesep}{1.5pt}
\setlength{\belowrulesep}{1.5pt}
\newcommand{\ourscell}[1]{\cellcolor{gray!10}#1}

\begin{adjustbox}{max width=\textwidth}
\begin{tabular}{
@{}
l l
c c c c
c
@{}
}
\toprule

&
&
\multicolumn{4}{c}{Generation}
& \multicolumn{1}{c}{Interaction}
\\

\cmidrule(lr){3-6}
\cmidrule(lr){7-7}

Setting
& Method
& FID$\downarrow$
& R@1/2/3$\uparrow$
& MM-Dist$\downarrow$
& Div. (GT)
& Contact F1$\uparrow$
\\
\midrule

\multirow{4}{*}{\shortstack[l]{1 Object\\(InterAct)}}
& CHOIS
& 0.2636
& 55.81/76.48/85.62
& 5.4210
& \textbf{1.3509} (1.3539)
& 26.09
\\

& MDM
& \textbf{0.1834}
& 24.31/68.35/79.27
& 8.7214
& 1.3593 (1.3539)
& 16.07
\\

& Dex2HOI
& 0.8662
& 50.83/71.34/81.30
& 5.9722
& 1.0300 (1.3539)
& 9.20
\\

& \ourscell{\textbf{Ours}}
& \ourscell{0.4871}
& \ourscell{\textbf{57.18/77.18/86.04}}
& \ourscell{\textbf{5.4181}}
& \ourscell{1.2207 (1.3539)}
& \ourscell{\textbf{28.11}}
\\

\specialrule{1.0pt}{2pt}{2pt}

\multirow{3}{*}{\shortstack[l]{2 Objects\\(HIMO)}}
& HIMO
& 0.3114
& 29.89/48.51/60.74
& 5.3877
& 1.3179 (1.4013)
& 12.73
\\

& Dex2HOI
& 0.4389
& 33.32/54.24/65.70
& 4.1991
& 1.2794 (1.4013)
& 43.03
\\

& \ourscell{\textbf{Ours}}
& \ourscell{\textbf{0.2253}}
& \ourscell{\textbf{35.55/58.89/69.34}}
& \ourscell{\textbf{4.0410}}
& \ourscell{\textbf{1.3519} (1.4013)}
& \ourscell{\textbf{44.95}}
\\

\specialrule{1.0pt}{2pt}{2pt}

\multirow{2}{*}{\shortstack[l]{3 Objects\\(HIMO)}}
& HIMO
& 0.3557
& 27.87/45.78/59.31
& 4.2359
& \textbf{1.3621} (1.3805)
& 12.49
\\

& \ourscell{\textbf{Ours}}
& \ourscell{\textbf{0.3074}}
& \ourscell{\textbf{29.34/48.09/61.28}}
& \ourscell{\textbf{4.1501}}
& \ourscell{1.3438 (1.3805)}
& \ourscell{\textbf{41.42}}
\\

\specialrule{1.0pt}{2pt}{2pt}

\multirow{3}{*}{\shortstack[l]{Ablation\\(HUMOTO)}}
& w/o Lifted Data
& 0.2475
& 11.64/22.50/32.19
& \textbf{3.6145}
& 6.3578 (6.1040)
& 32.22
\\

& w/o Pretraining
& 0.2616
& 11.88/20.86/30.08
& 3.7236
& 6.1744 (6.1040)
& 31.93
\\

& \ourscell{\textbf{Ours}}
& \ourscell{\textbf{0.2378}}
& \ourscell{\textbf{13.12/22.81/32.34}}
& \ourscell{3.6384}
& \ourscell{6.3291 (6.1040)}
& \ourscell{\textbf{34.91}}
\\

\bottomrule
\end{tabular}
\end{adjustbox}
\end{table*}

We evaluate BUMP on single- and multi-object HOI generation, followed by
quantitative comparisons and ablations of the proposed components.

\subsection{Datasets}
\label{sec:exp_datasets}

\noindent\textbf{Benchmark evaluation.}
For single-object interaction, we use InterAct, while for multi-object
interaction we use the two- and three-object subsets of HIMO as separate
evaluation settings.
Within each setting, BUMP and all compared methods are trained using the same
benchmark-specific training data and evaluated under the same split and
evaluation protocol.
The heterogeneous corpus introduced in Sec.~\ref{sec:data} is not used for
these apples-to-apples benchmark comparisons.

\noindent\textbf{Heterogeneous-supervision evaluation.}
We separately evaluate the effect of the heterogeneous training corpus from
Sec.~\ref{sec:data}, which combines human-only motion, captured full-body HOI,
and lifted manipulation data.
For these ablations, we keep the evaluation set fixed to the held-out HUMOTO
split and vary only the available training supervision.
The complete corpus contains 57,434 human-only training windows and 65,237
HOI training windows, including 13,545 lifted TACO and OakInk2 windows.

\begin{figure*}[t]
    \centering

    \includegraphics[width=\textwidth]{figures/qualitative_comparison.pdf}
    \caption{
        \textbf{Qualitative comparison across different numbers of interacting objects.}
        We compare representative generations on InterAct and HIMO using temporal snapshots and hand-object close-ups.
    }
    \label{fig:qualitative_comparison}

    \vspace{1em}

    \includegraphics[width=\textwidth]{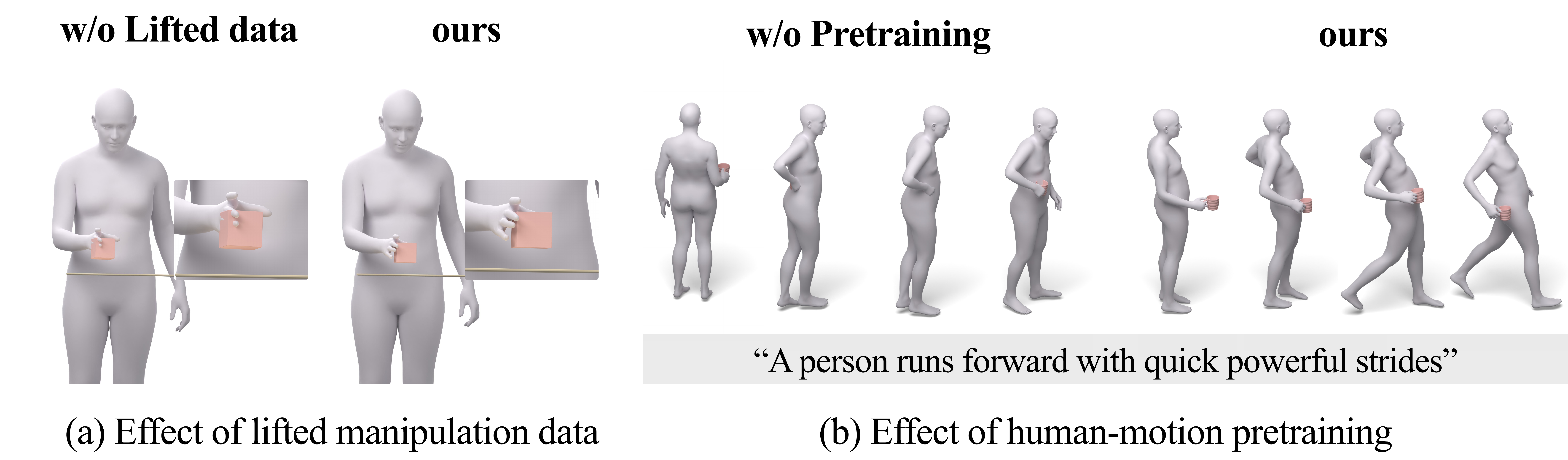}

    \caption{
        \textbf{Qualitative ablation of heterogeneous supervision.}
    }
    \label{fig:qualitative_ablation}
\end{figure*}

\subsection{Evaluation Metrics}
\label{sec:metrics}

We evaluate both the quality of the generated motion distribution and the
consistency of the resulting human-object interaction.
All metrics are computed independently for each benchmark setting; therefore,
scores are compared only among methods evaluated on the same dataset and
object-count.

\noindent\textbf{Generation quality and text alignment.}
Following the standard text-to-motion evaluation protocol~\citep{Guo2022HumanML3D},
we report Fr\'echet Inception Distance (FID), R-Precision (R@1/2/3),
Multimodal Distance (MM-Dist), and Diversity.
FID measures the discrepancy between the feature distributions of generated
and ground-truth motions, with lower values indicating better distributional
agreement.
R-Precision evaluates text-motion retrieval accuracy, while MM-Dist measures
the distance between matched text and motion embeddings.
Diversity measures variation among generated motions.
Because larger diversity is not necessarily better, we additionally report
the ground-truth diversity in parentheses and interpret generated values by
their proximity to this reference.

\noindent\textbf{Interaction quality.}
We evaluate hand-object interaction consistency using pairwise Contact F1.
For each frame, contact is determined separately for every hand-object pair
using a 2\,cm proximity threshold.
Contact F1 is computed over these hand-object-frame assignments, such that a
prediction is correct only when both the contacting hand and the contacted
object match the ground truth.
We report Contact F1 as a percentage.

\subsection{Results}
\label{sec:results}

\noindent\textbf{Quantitative comparisons.}
Table~\ref{tab:main_baselines} compares \modelname{} with prior methods across
single- and multi-object interaction settings.
\modelname{} achieves competitive generation quality on InterAct while
showing improved hand-object contact consistency.
The gains become more evident in the multi-object HIMO settings, where
\modelname{} maintains strong generation quality while substantially
improving Contact F1 over existing methods.
These results support our factorized formulation for coordinated
multi-object interaction generation.

\noindent\textbf{Qualitative comparisons.}
Figure~\ref{fig:qualitative_comparison} shows representative generations
across the single-, two-, and three-object settings.
\modelname{} produces coordinated full-body and object motion while maintaining the
hand-object relationships required by each interaction.
In multi-object examples, the generated motions additionally preserve the
coordination among simultaneously manipulated objects.

\subsection{Ablation Study}
\label{sec:ablation}

\noindent\textbf{Lifted manipulation data.}
We first study the effect of augmenting full-body HOI training with the
lifted TACO and OakInk2 sequences.
As shown in Table~\ref{tab:main_baselines}, removing the lifted data degrades
both generation quality and Contact F1 on the held-out HUMOTO evaluation.
This suggests that the additional hand-centric manipulation data provides
useful supervision for learning detailed multi-object interaction patterns.
Qualitatively, Fig.~\ref{fig:qualitative_ablation}(a) further shows improved
hand and finger coordination on a held-out GRAB sample.

\noindent\textbf{Human-motion pretraining.}
We next evaluate the contribution of human-only motion pretraining.
Pretraining improves FID, text alignment, and Contact F1, while also
increasing the diversity of the generated motions.
Interestingly, the resulting Diversity is higher than that of the held-out
HUMOTO distribution, suggesting that the human-motion prior broadens the
motion space beyond the relatively limited dynamics present in the HOI
training data.
As shown in Fig.~\ref{fig:qualitative_ablation}(b) on a held-out HUMOTO scene, this effect is particularly visible for prompts requiring larger and more dynamic full-body motion.

\section{Discussion and Conclusion}

We presented \modelname{}, a text-driven multi-object HOI framework that
separates intentional motion from physical consequences.
\modelname{} generates the human and a small number of manipulated driver
objects, while responder object motion is obtained through physics simulation.
We further leverage heterogeneous supervision from human-only, full-body HOI,
and lifted manipulation data, and introduce outcome-conditioned editing that
modifies the causal driver object trajectory from a desired responder object behavior.

\noindent\textbf{Limitations.}
Our current model supports at most two learned driver objects and assumes
rigid-body responder objects with prescribed human and driver object trajectories during
simulation.
The editing process is local and contact-driven, and part of our training data
relies on fitted full-body motion from hand-centric observations.
Extending the framework to more driver objects, richer physical interactions, and
long-horizon editing remains future work.

\bibliography{references/references}

\begin{thebibliography}{37}
\providecommand{\natexlab}[1]{#1}
\providecommand{\url}[1]{\texttt{#1}}
\expandafter\ifx\csname urlstyle\endcsname\relax
  \providecommand{\doi}[1]{doi: #1}\else
  \providecommand{\doi}{doi: \begingroup \urlstyle{rm}\Url}\fi

\bibitem[Bhatnagar et~al.(2022)Bhatnagar, Xie, Petrov, Sminchisescu, Theobalt, and Pons-Moll]{BEHAVE}
Bharat~Lal Bhatnagar, Xianghui Xie, Ilya~A Petrov, Cristian Sminchisescu, Christian Theobalt, and Gerard Pons-Moll.
\newblock Behave: Dataset and method for tracking human object interactions.
\newblock In \emph{2022 IEEE/CVF Conference on Computer Vision and Pattern Recognition (CVPR)}, pp.\  15914--15925. IEEE, 2022.

\bibitem[Deng et~al.(2021)Deng, Xu, Wu, Chen, and Jia]{3DAffordanceNet}
Shengheng Deng, Xun Xu, Chaozheng Wu, Ke~Chen, and Kui Jia.
\newblock 3d affordancenet: A benchmark for visual object affordance understanding.
\newblock In \emph{2021 IEEE/CVF Conference on Computer Vision and Pattern Recognition (CVPR)}, pp.\  1778--1787. IEEE, 2021.

\bibitem[Di et~al.(2026)Di, Li, Wang, Liu, Gao, and Ji]{PAMotion}
Yan Di, Yuheng Li, Yaoxing Wang, Mengge Liu, Shan Gao, and Xiangyang Ji.
\newblock Pamotion: Physics-aware motion generation for full-body interaction with multiple objects.
\newblock In \emph{Proceedings of the IEEE/CVF Conference on Computer Vision and Pattern Recognition (CVPR)}, pp.\  30661--30674, June 2026.

\bibitem[Fan et~al.(2023)Fan, Taheri, Tzionas, Kocabas, Kaufmann, Black, and Hilliges]{ARCTIC}
Zicong Fan, Omid Taheri, Dimitrios Tzionas, Muhammed Kocabas, Manuel Kaufmann, Michael~J Black, and Otmar Hilliges.
\newblock Arctic: A dataset for dexterous bimanual hand-object manipulation.
\newblock In \emph{2023 IEEE/CVF Conference on Computer Vision and Pattern Recognition (CVPR)}, pp.\  12943--12954. IEEE, 2023.

\bibitem[Guo et~al.(2022)Guo, Zou, Zuo, Wang, Ji, Li, and Cheng]{Guo2022HumanML3D}
Chuan Guo, Shihao Zou, Xinxin Zuo, Sen Wang, Wei Ji, Xingyu Li, and Li~Cheng.
\newblock Generating diverse and natural 3d human motions from text.
\newblock In \emph{2022 IEEE/CVF Conference on Computer Vision and Pattern Recognition (CVPR)}, pp.\  5142--5151. IEEE, 2022.

\bibitem[Harvey et~al.(2020)Harvey, Yurick, Nowrouzezahrai, and Pal]{LAFAN1}
F{\'e}lix~G Harvey, Mike Yurick, Derek Nowrouzezahrai, and Christopher Pal.
\newblock Robust motion in-betweening.
\newblock \emph{ACM Transactions on Graphics (TOG)}, 39\penalty0 (4):\penalty0 60--1, 2020.

\bibitem[Huang et~al.(2024)Huang, Taheri, Black, and Tzionas]{InterCap}
Yinghao Huang, Omid Taheri, Michael~J Black, and Dimitrios Tzionas.
\newblock Intercap: Joint markerless 3d tracking of humans and objects in interaction from multi-view rgb-d images.
\newblock \emph{International Journal of Computer Vision}, 132\penalty0 (7):\penalty0 2551--2566, 2024.

\bibitem[Kim et~al.(2024)Kim, Han, Kwon, and Joo]{ComA}
Hyeonwoo Kim, Sookwan Han, Patrick Kwon, and Hanbyul Joo.
\newblock Beyond the contact: Discovering comprehensive affordance for 3d objects from pre-trained 2d diffusion models.
\newblock In \emph{European Conference on Computer Vision}, pp.\  400--419. Springer, 2024.

\bibitem[Kim et~al.(2025{\natexlab{a}})Kim, Baik, and Joo]{DAViD}
Hyeonwoo Kim, Sangwon Baik, and Hanbyul Joo.
\newblock David: Modeling dynamic affordance of 3d objects using pre-trained video diffusion models.
\newblock In \emph{2025 IEEE/CVF International Conference on Computer Vision (ICCV)}, pp.\  10330--10341. IEEE, 2025{\natexlab{a}}.

\bibitem[Kim et~al.(2025{\natexlab{b}})Kim, Kim, Na, and Joo]{ParaHome}
Jeonghwan Kim, Jisoo Kim, Jeonghyeon Na, and Hanbyul Joo.
\newblock Parahome: Parameterizing everyday home activities towards 3d generative modeling of human-object interactions.
\newblock In \emph{2025 IEEE/CVF Conference on Computer Vision and Pattern Recognition (CVPR)}, pp.\  1816--1828. IEEE, 2025{\natexlab{b}}.

\bibitem[Li et~al.(2023)Li, Wu, and Liu]{OMOMO}
Jiaman Li, Jiajun Wu, and C~Karen Liu.
\newblock Object motion guided human motion synthesis.
\newblock \emph{ACM Transactions on Graphics (TOG)}, 42\penalty0 (6):\penalty0 1--11, 2023.

\bibitem[Li et~al.(2024)Li, Clegg, Mottaghi, Wu, Puig, and Liu]{CHOIS}
Jiaman Li, Alexander Clegg, Roozbeh Mottaghi, Jiajun Wu, Xavier Puig, and C~Karen Liu.
\newblock Controllable human-object interaction synthesis.
\newblock In \emph{European Conference on Computer Vision}, pp.\  54--72. Springer, 2024.

\bibitem[Li et~al.(2021)Li, Yang, Ross, and Kanazawa]{AISTPP}
Ruilong Li, Shan Yang, David~A. Ross, and Angjoo Kanazawa.
\newblock Ai choreographer: Music conditioned 3d dance generation with aist++, 2021.
\newblock URL \url{https://arxiv.org/abs/2101.08779}.

\bibitem[Liu et~al.(2024{\natexlab{a}})Liu, Zhu, Becherini, Peng, Su, Zhou, Zhe, Iwamoto, Zheng, and Black]{BEAT2}
Haiyang Liu, Zihao Zhu, Giorgio Becherini, Yichen Peng, Mingyang Su, You Zhou, Xuefei Zhe, Naoya Iwamoto, Bo~Zheng, and Michael~J Black.
\newblock Emage: Towards unified holistic co-speech gesture generation via expressive masked audio gesture modeling.
\newblock In \emph{2024 IEEE/CVF Conference on Computer Vision and Pattern Recognition (CVPR)}, pp.\  1144--1154. IEEE, 2024{\natexlab{a}}.

\bibitem[Liu et~al.(2024{\natexlab{b}})Liu, Yang, Si, Liu, Li, Zhang, Liu, and Yi]{TACO}
Yun Liu, Haolin Yang, Xu~Si, Ling Liu, Zipeng Li, Yuxiang Zhang, Yebin Liu, and Li~Yi.
\newblock Taco: Benchmarking generalizable bimanual tool-action-object understanding.
\newblock In \emph{2024 IEEE/CVF Conference on Computer Vision and Pattern Recognition (CVPR)}, pp.\  21740--21751. IEEE, 2024{\natexlab{b}}.

\bibitem[Lu et~al.(2025)Lu, Huang, Bhattacharya, Huang, and Zhou]{lu2025humoto}
Jiaxin Lu, Chun-Hao~Paul Huang, Uttaran Bhattacharya, Qixing Huang, and Yi~Zhou.
\newblock Humoto: A 4d dataset of mocap human object interactions.
\newblock In \emph{2025 IEEE/CVF International Conference on Computer Vision (ICCV)}, pp.\  10886--10897. IEEE, 2025.

\bibitem[Lv et~al.(2024)Lv, Xu, Yan, Jin, Xu, Wu, Liu, Li, Bi, Zeng, et~al.]{HIMO}
Xintao Lv, Liang Xu, Yichao Yan, Xin Jin, Congsheng Xu, Shuwen Wu, Yifan Liu, Lincheng Li, Mengxiao Bi, Wenjun Zeng, et~al.
\newblock Himo: A new benchmark for full-body human interacting with multiple objects.
\newblock In \emph{European Conference on Computer Vision}, pp.\  300--318. Springer, 2024.

\bibitem[Peng et~al.(2025)Peng, Xie, Wu, Jampani, Sun, and Jiang]{HOIDiff}
Xiaogang Peng, Yiming Xie, Zizhao Wu, Varun Jampani, Deqing Sun, and Huaizu Jiang.
\newblock Hoi-diff: Text-driven synthesis of 3d human-object interactions using diffusion models.
\newblock In \emph{2025 IEEE/CVF Conference on Computer Vision and Pattern Recognition Workshops (CVPRW)}, pp.\  2869--2879. IEEE, 2025.

\bibitem[Pratikaki et~al.(2026)Pratikaki, Ruiz-Ponce, Deng, Zafeiriou, and Potamias]{Dex2HOI}
Chrysa Pratikaki, Pablo Ruiz-Ponce, Jiankang Deng, Stefanos Zafeiriou, and Rolandos~Alexandros Potamias.
\newblock Dex2hoi: Dexterous bimanual two-object interaction generation, 2026.
\newblock URL \url{https://arxiv.org/abs/2605.30444}.

\bibitem[Prokudin et~al.(2019)Prokudin, Lassner, and Romero]{prokudin2019efficient}
Sergey Prokudin, Christoph Lassner, and Javier Romero.
\newblock Efficient learning on point clouds with basis point sets.
\newblock In \emph{Proceedings of the IEEE International Conference on Computer Vision}, pp.\  4332--4341, 2019.

\bibitem[Rempe et~al.(2026)Rempe, Petrovich, Yuan, Zhang, Peng, Jiang, Wang, Iqbal, Minor, de~Ruyter, et~al.]{rempe2026kimodo}
Davis Rempe, Mathis Petrovich, Ye~Yuan, Haotian Zhang, Xue~Bin Peng, Yifeng Jiang, Tingwu Wang, Umar Iqbal, David Minor, Michael de~Ruyter, et~al.
\newblock Kimodo: Scaling controllable human motion generation.
\newblock \emph{arXiv preprint arXiv:2603.15546}, 2026.

\bibitem[Taheri et~al.(2020)Taheri, Ghorbani, Black, and Tzionas]{GRAB}
Omid Taheri, Nima Ghorbani, Michael~J Black, and Dimitrios Tzionas.
\newblock Grab: A dataset of whole-body human grasping of objects.
\newblock In \emph{European conference on computer vision}, pp.\  581--600. Springer, 2020.

\bibitem[Todorov et~al.(2012)Todorov, Erez, and Tassa]{todorov2012mujoco}
Emanuel Todorov, Tom Erez, and Yuval Tassa.
\newblock Mujoco: A physics engine for model-based control.
\newblock In \emph{2012 IEEE/RSJ international conference on intelligent robots and systems}, pp.\  5026--5033. IEEE, 2012.

\bibitem[Tripathi et~al.(2023)Tripathi, Chatterjee, Passy, Yi, Tzionas, and Black]{DECO}
Shashank Tripathi, Agniv Chatterjee, Jean-Claude Passy, Hongwei Yi, Dimitrios Tzionas, and Michael~J Black.
\newblock Deco: Dense estimation of 3d human-scene contact in the wild.
\newblock In \emph{2023 IEEE/CVF International Conference on Computer Vision (ICCV)}, pp.\  7967--7979. IEEE, 2023.

\bibitem[Wang et~al.(2024)Wang, Chen, Jia, Li, Zhang, Zhang, Liu, Zhu, Liang, and Huang]{AffordMotion}
Zan Wang, Yixin Chen, Baoxiong Jia, Puhao Li, Jinlu Zhang, Jingze Zhang, Tengyu Liu, Yixin Zhu, Wei Liang, and Siyuan Huang.
\newblock Move as you say, interact as you can: Language-guided human motion generation with scene affordance.
\newblock In \emph{2024 IEEE/CVF Conference on Computer Vision and Pattern Recognition (CVPR)}, pp.\  433--444. IEEE, 2024.

\bibitem[Wasserberger et~al.(2022)Wasserberger, Brady, Besky, Jones, and Boddy]{OBP}
KW~Wasserberger, AC~Brady, DM~Besky, BR~Jones, and KJ~Boddy.
\newblock The openbiomechanics project: The open source initiative for anonymized, elite-level athletic motion capture data.
\newblock 2022.

\bibitem[Wei et~al.(2022)Wei, Liu, Ling, and Su]{wei2022coacd}
Xinyue Wei, Minghua Liu, Zhan Ling, and Hao Su.
\newblock Approximate convex decomposition for 3d meshes with collision-aware concavity and tree search.
\newblock \emph{ACM Transactions on Graphics (TOG)}, 41\penalty0 (4):\penalty0 1--18, 2022.

\bibitem[Wu et~al.(2025)Wu, Li, Xu, and Liu]{HOIfHLI}
Zhen Wu, Jiaman Li, Pei Xu, and C~Karen Liu.
\newblock Human-object interaction from human-level instructions.
\newblock In \emph{2025 IEEE/CVF International Conference on Computer Vision (ICCV)}, pp.\  11176--11186. IEEE, 2025.

\bibitem[Xu et~al.(2023)Xu, Li, Wang, and Gui]{InterDiff}
Sirui Xu, Zhengyuan Li, Yu-Xiong Wang, and Liang-Yan Gui.
\newblock Interdiff: Generating 3d human-object interactions with physics-informed diffusion.
\newblock In \emph{2023 IEEE/CVF International Conference on Computer Vision (ICCV)}, pp.\  14882--14894. IEEE, 2023.

\bibitem[Xu et~al.(2024)Xu, Wang, Wang, and Gui]{InterDreamer}
Sirui Xu, Ziyin Wang, Yu-Xiong Wang, and Liang-Yan Gui.
\newblock Interdreamer: Zero-shot text to 3d dynamic human-object interaction.
\newblock \emph{Advances in Neural Information Processing Systems}, 37:\penalty0 52858--52890, 2024.

\bibitem[Xu et~al.(2025)Xu, Li, Zhang, Xu, Long, Wang, Lu, Dong, Jiang, Gupta, et~al.]{InterAct}
Sirui Xu, Dongting Li, Yucheng Zhang, Xiyan Xu, Qi~Long, Ziyin Wang, Yunzhi Lu, Shuchang Dong, Hezi Jiang, Akshat Gupta, et~al.
\newblock Interact: Advancing large-scale versatile 3d human-object interaction generation.
\newblock In \emph{2025 IEEE/CVF Conference on Computer Vision and Pattern Recognition (CVPR)}, pp.\  7048--7060. IEEE, 2025.

\bibitem[Yang et~al.(2023)Yang, Zhai, Luo, Cao, Luo, and Zha]{IAGNet}
Yuhang Yang, Wei Zhai, Hongchen Luo, Yang Cao, Jiebo Luo, and Zheng-Jun Zha.
\newblock Grounding 3d object affordance from 2d interactions in images.
\newblock In \emph{2023 IEEE/CVF International Conference on Computer Vision (ICCV)}, pp.\  10871--10881. IEEE, 2023.

\bibitem[Yang et~al.(2024)Yang, Zhai, Luo, Cao, and Zha]{LEMON}
Yuhang Yang, Wei Zhai, Hongchen Luo, Yang Cao, and Zheng-Jun Zha.
\newblock Lemon: Learning 3d human-object interaction relation from 2d images.
\newblock In \emph{2024 IEEE/CVF Conference on Computer Vision and Pattern Recognition (CVPR)}, pp.\  16284--16295. IEEE, 2024.

\bibitem[Yi et~al.(2024)Yi, Thies, Black, Peng, and Rempe]{TesMo}
Hongwei Yi, Justus Thies, Michael~J Black, Xue~Bin Peng, and Davis Rempe.
\newblock Generating human interaction motions in scenes with text control.
\newblock In \emph{European Conference on Computer Vision}, pp.\  246--263. Springer, 2024.

\bibitem[Zhan et~al.(2024)Zhan, Yang, Zhao, Mao, Xu, Lin, Li, and Lu]{OakInk2}
Xinyu Zhan, Lixin Yang, Yifei Zhao, Kangrui Mao, Hanlin Xu, Zenan Lin, Kailin Li, and Cewu Lu.
\newblock Oakink2: A dataset of bimanual hands-object manipulation in complex task completion.
\newblock In \emph{2024 IEEE/CVF Conference on Computer Vision and Pattern Recognition (CVPR)}, pp.\  445--456. IEEE, 2024.

\bibitem[Zhang et~al.(2025)Zhang, Lin, Zeng, Wu, Lu, Fu, Cai, Zhang, Wang, and Zhang]{zhang2025motion}
Yuhong Zhang, Jing Lin, Ailing Zeng, Guanlin Wu, Shunlin Lu, Yurong Fu, Yuanhao Cai, Ruimao Zhang, Haoqian Wang, and Lei Zhang.
\newblock Motion-x++: A large-scale multimodal 3d whole-body human motion dataset.
\newblock \emph{arXiv preprint arXiv:2501.05098}, 2025.

\bibitem[Zhou et~al.(2019)Zhou, Barnes, Lu, Yang, and Li]{zhou2019continuity}
Yi~Zhou, Connelly Barnes, Jingwan Lu, Jimei Yang, and Hao Li.
\newblock On the continuity of rotation representations in neural networks.
\newblock In \emph{2019 IEEE/CVF Conference on Computer Vision and Pattern Recognition (CVPR)}, pp.\  5738--5746. IEEE, 2019.

\end{thebibliography}
\bibliographystyle{iclr2027_conference}

\end{document}